# Intercausal Reasoning with Uninstantiated Ancestor Nodes

Marek J. Druzdzel
Carnegie Mellon University
Department of Engineering and Public Policy
Pittsburgh, PA 15213
marek+@cmu.edu

Max Henrion
Rockwell International Science Center
Palo Alto Laboratory
444 High Street, Suite 400
Palo Alto, CA 94301
henrion@camis.stanford.edu

## Abstract

Intercausal reasoning is a common inference pattern involving probabilistic dependence of causes of an observed common effect. The sign of this dependence is captured by a qualitative property called *product synergy*. The current definition of product synergy is insufficient for intercausal reasoning where there are additional uninstantiated causes of the common effect. We propose a new definition of product synergy and prove its adequacy for intercausal reasoning with direct and indirect evidence for the common effect. The new definition is based on a new property *matrix half positive semi-definiteness*, a weakened form of matrix positive semi-definiteness.

## 1  INTRODUCTION

Intercausal reasoning is a common inference pattern involving probabilistic dependence of causes of an observed common effect. The most common form of intercausal reasoning is "explaining away" (Henrion, 1986; Pearl, 1988), which is when given an observed effect and increase in probability of one cause, all other causes of that effect become less likely. For example, even though the use of fertilizer early in the spring and the weather throughout the growing season can be assumed to be probabilistically independent, once we know that the crop was extremely good, this independence vanishes. Upon having heard that the weather was extraordinarily good, we find that the likelihood that an efficient fertilizer had been used early in the spring diminishes — good weather "explains away" the fertilizer. Although explaining away appears to be the most common pattern of intercausal reasoning, the reverse is also possible, i.e., observing one cause can make other causes more likely. We call both types of reasoning "intercausal," although strictly speaking it is not necessary that the variables involved are in causal relationships with one another and our subsequent analysis captures probabilistic rather than causal conditions. In fact, two variables $a$ and $b$ will be in an "intercausal relationship" given a third variable $c$ if they are independent conditional on a set of variables $\Psi$, but dependent conditional on every set $\Phi$ such that $c \in \Phi$. This is captured by the graphical structure of a Bayesian belief network in which $a$ and $b$ are direct predecessors of $c$ and there is no arc between $a$ and $b$. The applications of intercausal reasoning include algorithms for belief updating in qualitative probabilistic networks (Druzdzel & Henrion, 1993; Henrion & Druzdzel, 1991), approximate search-based algorithms for BBNs (Henrion, 1991), and automatic generation of explanations of probabilistic reasoning in decision support systems (Druzdzel, 1993).

Intercausal reasoning has been captured formally by a qualitative property called *product synergy* (Henrion & Druzdzel, 1991; Wellman & Henrion, 1991). The sign of the product synergy determines the sign of the intercausal influence. Previous work on intercausal reasoning, and product synergy in particular, concentrated on situations where all irrelevant ancestors of the common effect were assumed to be instantiated. In this paper we propose a new definition of product synergy that enables performing intercausal reasoning in arbitrary belief networks. We prove that the new definition is sufficient for intercausal reasoning with the common effect observed and is also sufficient for intercausal reasoning with indirect support when the common effect variable is binary.

The influence of indirect evidential support on intercausal reasoning in the binary common effect case has been studied before by Wellman and Henrion (1991) and by Agosta (1991). Our exposition deals with the general case including uninstantiated predecessors of the common effect variable and, therefore, advances insight into intercausal reasoning beyond what has been presented in those papers. Another difference between this and Wellman and Henrion's exposition is that here we provide more insight into the functional dependences between nodes in intercausal reasoning. We improve their theorem listing the conditions for intercausal reasoning with indirect evidential support. We generalize Agosta's analysis of intercausal reasoning from the binary case. Our analysis of Conditional Inter-Causally Independent (CICI) node distributions shows that there is a large class of relations for which non-trivial evidential support can leave their direct ancestors independent.

All random variables that we deal with in this paper are multiply valued, discrete variables, such as those represented by nodes of a Bayesian belief network. We make this assumption for the reasons of convenience in mathematical derivations and proofs.

Following Wellman (1990), we will assume that all conditional probability terms are well defined and those that appear in the denominators are non-zero. This assumption is easily relaxed at the cost of explicatory complexity.

Lower case letters (e.g., $x$) will stand for random variables, indexed lower-case letters (e.g., $x_i$) will usually denote their outcomes. In case of binary random variables, the two outcomes will be denoted by upper case (e.g., the two outcomes of a variable $c$ will be denoted by $C$ and $\overline{C}$). Outcomes of random variables are ordered from the highest to the lowest value. And so, for a random variable $a$, $\forall_{i<j}\ [a_i \geq a_j]$. For binary variables $C > \overline{C}$, or **true**>**false**. Indexed lower case letter $n$, such as $n_a$ denotes the number of outcomes of a variable $a$.

We will use bold upper-case letters (e.g., **M**) and bold lower-case letters (e.g., **x**) for matrices and vectors respectively. Elements of matrices will be doubly indexed upper-case letters (e.g., $M_{ij}$).

The remainder of this paper is structured as follows. Section 2 reviews the elementary qualitative properties of probabilistic interactions, as captured in qualitative probabilistic networks. Section 3 demonstrates the problem of sensitivity of the previous definition of product synergy to the probability distribution over the values of uninstantiated direct ancestors of the common effect node. Section 4 proposes a new definition of product synergy that is provably sufficient and necessary for intercausal reasoning and studies the properties of intercausal reasoning when the evidential support for the common effect is direct and indirect. We discuss intercausal reasoning in Noisy-OR gates in Section 5. Detailed proofs of all theorems can be found in the appendix.

## 2 QUALITATIVE PROBABILISTIC NETWORKS

Qualitative probabilistic networks (QPNs) (Wellman, 1990) are an abstraction of Bayesian belief networks replacing numerical relations by specification of qualitative properties. So far, three qualitative properties of probability distributions have been formalized: qualitative influence, additive synergy, and product synergy. Since we will refer to them later in the paper, we reproduce the definitions of these properties here after (Wellman & Henrion, 1991).

**Definition 1 (qualitative influence)** *We say that $a$ positively influences $c$, written $S^+(a,c)$, iff for all values $a_1 > a_2$, $c_0$, and $x$,*

$$\Pr(c \geq c_0|a_1 x) \geq \Pr(c \geq c_0|a_2 x).$$

This definition expresses the fact that increasing the value of $a$, makes higher values of $c$ more probable. *Negative qualitative influence*, $S^-$, and *zero qualitative influence*, $S^0$, are defined analogously by substituting $\geq$ by $\leq$ and $=$ respectively.

**Definition 2 (additive synergy)** *Variables $a$ and $b$ exhibit positive additive synergy with respect to variable $c$, written $Y^+(\{a,b\},c)$, if for all $a_1 > a_2$, $b_1 > b_2$, $x$, and $c_0$,*

$$Pr(c \geq c_0|a_1 b_1 x) + Pr(c \geq c_0|a_2 b_2 x)$$
$$\geq Pr(c \geq c_0|a_1 b_2 x) + Pr(c \geq c_0|a_2 b_1 x).$$

The additive synergy is used with respect to two causes and a common effect. It captures the property that the joint influence of the two causes is greater than sum of their individual effects. *Negative additive synergy*, $Y^-$, and *zero additive synergy*, $Y^0$, are defined analogously by substituting $\geq$ by $\leq$ and $=$ respectively.

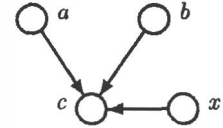

Figure 1: Intercausal reasoning between $a$ and $b$ with an additional predecessor variable $x$.

**Definition 3 (product synergy I)** *Let $a$, $b$, and $x$ be the predecessors of $c$ in a QPN (see Figure 1). Variables $a$ and $b$ exhibit negative product synergy with respect to a particular value $c_0$ of c, written $X^-(\{a,b\},c_0)$, if for all $a_1 > a_2$, $b_1 > b_2$, and $x$,*

$$Pr(c_0|a_1 b_1 x)Pr(c_0|a_2 b_2 x)$$
$$\leq Pr(c_0|a_1 b_2 x)Pr(c_0|a_2 b_1 x).$$

*Positive product synergy*, $X^+$, and *zero product synergy*, $X^0$, are defined analogously by substituting $\leq$ by $\geq$ and $=$ respectively. Note that product synergy is defined with respect to each outcome of the common effect $c$. There are, therefore, as many product synergies as there are outcomes in $c$. For a binary variable $c$, there are two product synergies, one for $C$ and one for $\overline{C}$. The practical implication of product synergy is that, under the specified circumstances, it forms a sufficient condition for explaining away.

## 3 UNINSTANTIATED ANCESTOR NODES

If $a$ and $x$ are both predecessors of $c$, with conditional probability distribution $Pr(c|ax)$, then the relation be-tween $a$ and $c$ depends on $x$. In other words, the probabilistic influence of one variable on another may depend on additional variables. Hence, the qualitative properties defined above contain the strong condition that they must hold for all possible instantiations of $x$. If the "irrelevant" node $x$ is uninstantiated, $x$ will not affect the signs of qualitative influence or additive synergy. But, surprisingly, it turns out that unobserved predecessors may affect the product synergy.

We will present an example showing that the presence of uninstantiated predecessor variables can affect the intercausal relation between other parents and explain informally the reasons for that effect. The example of a simple BBN with binary variables, the associated conditional probability distribution of the common effect node, and the resulting qualitative properties of the interaction between the variables, are given in Figure 2. The qualitative properties of the interaction among $a$, $b$, $c$, and $x$ are all well defined. In particular, product synergy I for $C$ observed is for $a$, $b$, and $x$ pairwise negative. Still, for some distributions of $x$, for example for $Pr(X) = 0.5$, the intercausal influence of $a$ on $b$ is positive (see Figure 3).

Quantitative conditional distribution:

| $Pr(C\|abx)$ | X | | $\overline{X}$ | |
|---|---|---|---|---|
| | B | $\overline{B}$ | B | $\overline{B}$ |
| A | 0.99 | 0.8 | 0.99 | 0.2 |
| $\overline{A}$ | 0.8 | 0.6 | 0.2 | 0.0 |

Qualitative properties:

$S^+(a,c)$    $X^-(\{a,b\},C)$
$S^+(b,c)$    $X^-(\{a,x\},C)$
$S^+(x,c)$    $X^-(\{b,x\},C)$

Figure 2: Example of the effect of an uninstantiated predecessor node $x$ on intercausal reasoning (see Figure 1). All pairwise product synergies for $C$ observed between $a$, $b$, and $x$ are negative and all influences of $a$, $b$, and $x$ on $c$ are positive.

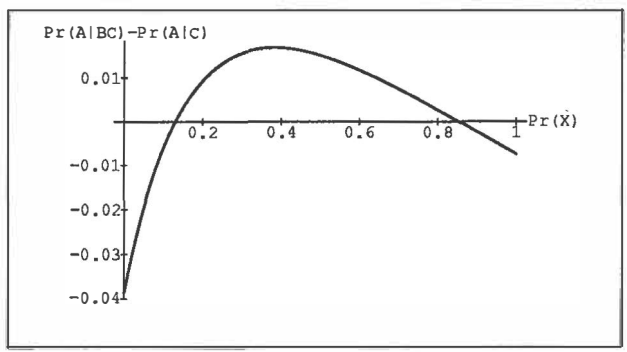

Figure 3: The intercausal interaction between $a$ and $b$ as a function of probability of $x$.

We propose the following explanation of this phenomenon. The sign of intercausal interaction between $a$ and $b$ is a function of the probability distribution of $x$. This function is not linear (it will become apparent in Section 4 that it is quadratic) and the fact that the function has the same sign at the extremes does not guarantee the same sign in all the points in between. In the example above, we are dealing with negative signs at the extremes (i.e., for $Pr(X) = 0$ and $Pr(X) = 1$) and a positive sign for some interval in between (see Figure 3).

## 4 PRODUCT SYNERGY II

A key objective for any qualitative property between two variables in a network is that this is invariant to the probability distribution of other neighboring nodes. This invariance allows for drawing conclusions that are valid regardless of the numerical values of probability distributions of the neighboring variables. As we have shown in the previous section, this does not apply to product synergy as previously defined. In this section, we propose a new definition of product synergy that will have this property. The new definition of product synergy is expressed in terms of a condition that we term *matrix half positive semi-definiteness*.

### 4.1 MATRIX HALF POSITIVE SEMI-DEFINITENESS

Half positive semi-definiteness is a weakened form of positive semi-definiteness (see for example (Strang, 1976)). A square $n \times n$ matrix $\mathbf{M}$ is positive semi-definite if and only if for any vector $\mathbf{x}$, $\mathbf{x}^T\mathbf{M}\mathbf{x} \geq 0$. $\mathbf{M}$ is half positive semi-definite if the above inequality holds for any non-negative vector $\mathbf{x}$.

**Definition 4 (non-negative matrix)** *A matrix is called non-negative (non-positive) if all its elements are non-negative (non-positive).*

**Definition 5 (half positive semi-definiteness)** *A square $n \times n$ matrix $\mathbf{M}$ is called half positive semi-definite (half negative semi-definite) if for any non-negative vector $\mathbf{x}$ consisting of $n$ elements $\mathbf{x}^T\mathbf{M}\mathbf{x} \geq 0$ ($\mathbf{x}^T\mathbf{M}\mathbf{x} \leq 0$).*

The following theorem addresses the problem of testing whether a given matrix is half positive semi-definite.

**Theorem 1 (half positive semi-definiteness)** *A sufficient condition for half positive semi-definiteness of a matrix is that it is a sum of a positive semi-definite and a non-negative matrix.*

It can be easily shown that the condition is also necessary for $2 \times 2$ matrices.

**Theorem 2 ($2 \times 2$ half positive semi-definiteness)**

*A necessary condition for half positive semi-definiteness of a 2 × 2 matrix is that it is a sum of a positive semi-definite and a non-negative matrix.*

We can prove this condition also for 3 × 3 matrices. We conjecture that this condition is true for $n \times n$ matrices, although so far we have not been able to find a general proof.

**Conjecture 1 (half positive semi-definiteness)**
*A sufficient and necessary condition for half positive semi-definiteness of a square matrix is that it is a sum of a positive semi-definite and a non-negative matrix.*

Given Theorem 1, we are still left with the problem of decomposing a $n \times n$ matrix into a sum of two matrices of which one is positive semi-definite and the other is non-negative. It can be easily shown that this decomposition is not unique. It seems that a practical procedure for determining whether a matrix is half positive semi-definite needs to be based on heuristic methods. It is easy to prove that half positive semi-definiteness necessitates $\forall_i \ [D_{ii} \leq 0]$ (consider a vector $x$ in which only $x_i$ is non-zero). The first test for any matrix is, therefore, whether the diagonal elements are non-negative. The heuristic methods might first check whether the matrix is positive semi-definite by studying its eigenvalues, pivots, or the determinants of its upper left submatrices. Another easy check is whether the matrix is non-negative. For any quadratic form, there exists an equivalent symmetric form, so the matrix will be non-negative if and only if all its diagonal elements are non-negative and $\forall_{ij} \ [D_{ij} + D_{ji} \geq 0]$, i.e., the sum of each pair of symmetric off-diagonal symmetric elements is non-negative. If both tests fail, one might try to decompose the matrix by subtracting from its elements positive numbers in such a way that it becomes positive semi-definite. The subtracted elements compose the non-negative matrix. As already indicated, this decomposition is not unique.

### 4.2 PRODUCT SYNERGY II

**Definition 6 (product synergy II)** *Let $a$, $b$, and $x$ be direct predecessors of $c$ in a QPN (see Figure 1). Let $n_x$ denote the number of possible values of $x$. Variables $a$ and $b$ exhibit negative product synergy with respect to a particular value $c_0$ of $c$, regardless of the distribution of $x$, written $X^-(\{a,b\}, c_0)$, if for all $a_1 > a_2$ and for all $b_1 > b_2$, a square $n_x \times n_x$ matrix $\mathbf{D}$ with elements*

$$D_{ij} = Pr(c_0|a_1b_1x_i)Pr(c_0|a_2b_2x_j)$$
$$\qquad - Pr(c_0|a_2b_1x_i)Pr(c_0|a_1b_2x_j).$$

*is half negative semi-definite. If $\mathbf{D}$ is half positive semi-definite, $a$ and $b$ exhibit positive product synergy written as $X^+(\{a,b\},c_0)$. If $\mathbf{D}$ is a zero matrix, $a$ and $b$ exhibit zero product synergy written as $X^0(\{a,b\},c_0)$.*

Note that although the definition of product synergy II covers the situation in which there is only one uninstantiated direct predecessor of $c$, it is easily extensible to the general case. If there are more than one uninstantiated direct predecessors, we can conceptually replace them by a single uninstantiated variable with the number of outcomes being the product of the number of outcomes of each variable separately. This is equivalent to rearranging the conditional distribution matrix of $c$.

Unless specified otherwise, in the remainder of this paper we will use the term *product synergy* meaning *product synergy II*. As product synergy I is a special case of product synergy II, we propose to adopt this convention in future references to this work.

### 4.3 INTERCAUSAL REASONING

The following theorem binds product synergy with intercausal reasoning in case when the common effect has been observed.

**Theorem 3 (intercausal reasoning)** *Let $a$, $b$, and $x$ be direct predecessors of $c$ such that $a$ and $b$ are conditionally independent. A sufficient and necessary condition for $S^-(a,b)$ on observation of $c_0$ is negative product synergy, $X^-(\{a,b\},c_0)$.*

### 4.4 INTERCAUSAL REASONING WITH INDIRECT EVIDENCE

The following theorem binds product synergy with intercausal reasoning in case of indirect support for a binary common effect.

**Theorem 4 (intercausal reasoning)** *Let $a$, $b$, and $x$ be direct predecessors of $c$, and $c$ be a direct predecessor of $d$ in a network. Let $c$ be binary. Let there be no direct links from $a$ or $b$ to $d$ (see Figure 4). Let $X^{\delta_1}(\{a,b\}, C)$, $X^{\delta_2}(\{a,b\}, \overline{C})$, $Y^{\delta_3}(\{a,b\}, c)$, and $S^{\delta_4}(c,d)$.*
*If $\delta_4 = +$ and $\delta_1 = \delta_3$, then $S^{\delta_1}(a,b)$ holds in the network with $D$ observed. If $\delta_4 = -$ and $\delta_2 \neq \delta_3$, then $S^{\delta_2}(a,b)$ holds in the network with $D$ observed.*

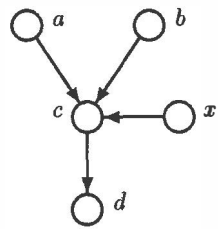

$X^{\delta_1}(\{a,b\}, C)$    $Y^{\delta_3}(\{a,b\}, c)$
$X^{\delta_2}(\{a,b\}, \overline{C})$    $S^{\delta_4}(c,d)$

Figure 4: Intercausal reasoning with indirect evidence. $d$ is observed, $x$ is uninstantiated.

Theorem 4 is an improvement on the theorem proposed by Wellman and Henrion (1991, Theorem 6), capturing additional conditions under which the sign of intercausal inference with indirect support can be resolved.

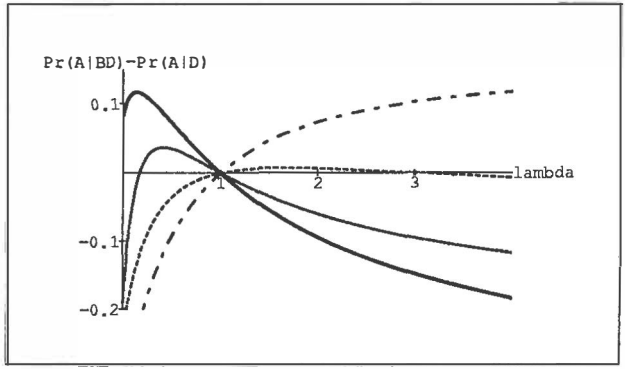

Figure 5: The intercausal interaction between $a$ and $b$ as a function of the evidential support for $c$ for various values of qualitative properties of interaction between $a$ and $b$.

Figure 5 shows the magnitude of intercausal interaction between $a$ and $b$ as a function of indirect evidential support for $C$, for different values of qualitative properties of interaction between $a$ and $b$. The strength of evidential support is expressed by $\lambda$, the likelihood ratio of the observed evidence ($\lambda = Pr(D|C)/Pr(D|\overline{C})$). The intercausal influence between $a$ and $b$ is, as expected, always zero for $\lambda = 1.0$ (no evidential support). $\lambda = 0$ corresponds to perfect evidence against $C$ (in other words, $\overline{C}$ is implied by $D$). $\lambda = \infty$ corresponds to perfect evidence for $c$ (in other words, $C$ is implied by $D$). In the proof of Theorem 4, we demonstrate that the interaction is quadratic in $\lambda$ and each of the curves has at most two zero points (one of these is a trivial zero point, for $\lambda = 1.0$). This result is in agreement with Agosta's (1991) finding that intercausal conditional independence in binary variables is possible at most at one state of evidence.

The product synergy and the additive synergy determine exactly the interval where the second zero point falls. The product synergies between $a$ and $b$ for $\overline{C}$ and $C$ determine whether the curve is above or below zero for $\lambda = 0$ and $\lambda = \infty$ respectively. The additive synergy helps to locate the second zero point of the curve. If the evidence is positive ($\lambda > 1.0$), and the additive synergy is equal to the positive product synergy, then the second zero point is for $\lambda < 1.0$. If the evidence is negative ($0 \leq \lambda < 1.0$), and the additive synergy is not equal to the negative product synergy, then the second zero point is for $\lambda > 1.0$.

## 5 NOISY-OR DISTRIBUTIONS

Noisy-OR gates (Pearl, 1988) are a common form of probabilistic interaction used in probabilistic models. It turns out that Noisy-OR gates are robust against the effect of uninstantiated predecessor variables discussed in Section 3. The conditional probability distribution of Noisy-OR gates always results in half negative semi-definite matrices used in the definition of product synergy and, effectively, the probability distribution of predecessor nodes never impacts intercausal reasoning.

### 5.1 UNINSTANTIATED PREDECESSOR VARIABLES

We will demonstrate the behavior of a leaky Noisy-OR gate $c$ with direct binary predecessors $a$, $b$, and $x$ (see Figure 1). Let $p$, $q$, and $r$, be the inhibitor probabilities (Pearl, 1988) for nodes $a$, $b$, and $x$ with respect to the node $c$ and $l$ be the leak probability. This determines the elements $D_{ij}$ of the matrix $\mathbf{D}$ (see Definition 6) to be

$$\begin{aligned}
D_{11} &= -(1-l)pq(1-r) \\
D_{12} &= -(1-l)q(p+(1-p)r) \\
D_{21} &= -(1-l)q(p-r) \\
D_{22} &= -(1-l)pq
\end{aligned}$$

It is easy to verify that $D_{11} \leq 0$ and $D_{22} \leq 0$. Also,

$$D_{12} + D_{21} = -(1-l)pq(2-r) \leq 0,$$

which shows that irrespective of the actual values of $p$, $q$, $r$, and $l$, a symmetric form of the matrix $\mathbf{D}$ is non-positive and, by Theorem 1, half positive semi-definite. Binary Noisy-OR gates will, therefore, always exhibit negative product synergy for the effect observed, regardless of presence or absence of uninstantiated predecessor variable $x$.

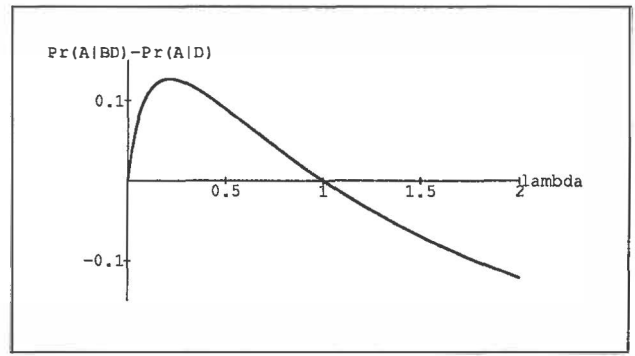

Figure 6: The intercausal interaction between $a$ and $b$ as a function of the evidential support for $c$ in a Noisy-OR gate.

### 5.2 INDIRECT EVIDENCE

As the product synergy given effect observed to be absent is equal to zero (i.e., for any Noisy-OR gate we have $X^0(\{a,b\},\overline{C})$), and the product synergy given effect observed is negative (i.e., for any Noisy-OR gate

$X^-(\{a,b\}, C))$, Equation 13 (see the proof of Theorem 4) reduces to

$$(\lambda - 1)\lambda \left| X^-(\{a,b\}, C) \right| \leq 0.$$

The two zero points of this expression with respect to $\lambda$ are for $\lambda = 0$ and $\lambda = 1$. We know that there are no other zero points (as shown in the proof of Theorem 4), and it follows that intercausal influences in Noisy-OR gates will always be negative for $\lambda > 1$ (positive evidence) and positive for $0 < \lambda < 1$ (negative evidence) (see Figure 6).

## 6  CONCLUSION

The previous definition of *product synergy* does not cover situations where there are additional uninstantiated causes of the common effect. We have introduced a new definition of product synergy and we proved its adequacy for intercausal reasoning with common effect directly observed and also intercausal reasoning with indirect evidential support when the common effect is binary. We introduced the term *matrix half positive semi-definiteness*, a weakened form of matrix positive semi-definiteness.

Intercausal reasoning is useful in qualitative schemes for reasoning under uncertainty and, because of its prevalence in commonsense reasoning, valuable for automatic generation of explanations of probabilistic reasoning. The new definition of product synergy allows for intercausal reasoning in arbitrary belief networks and directly supports both tasks. As probabilities are non-negative and, in many cases, the condition of matrix positive definiteness may be too strong, we suspect that the property of matrix half positive definiteness that we introduced in this paper will prove theoretically useful in qualitative analysis of probabilistic reasoning.

## APPENDIX: PROOFS

**Theorem 1 (half positive semi-definiteness)** *A sufficient condition for half positive semi-definiteness of a matrix is that it is a sum of a positive semi-definite and a non-negative matrix.*

**Proof:**  Let $\mathbf{M} = \mathbf{M}_1 + \mathbf{M}_2$, where $\mathbf{M}_1$ is positive semi-definite and $\mathbf{M}_2$ is non-negative. Since positive semi-definiteness holds for any vector $\mathbf{x}$, and in particular for a non-negative one, a positive semi-definite matrix $\mathbf{M}_1$ is also a half positive semi-definite. We have therefore, $\mathbf{x}^T \mathbf{M}_1 \mathbf{x} \geq 0$. Also, a non-negative matrix is half positive semi-definite, since any quadratic form with all non-negative elements cannot be negative. We have therefore that $\mathbf{x}^T \mathbf{M}_2 \mathbf{x} \geq 0$. Sum of two non-negative numbers is non-negative, therefore $\mathbf{x}^T \mathbf{M}_1 \mathbf{x} + \mathbf{x}^T \mathbf{M}_2 \mathbf{x} \geq 0$. By elementary matrix algebra

$$0 \leq \mathbf{x}^T \mathbf{M}_1 \mathbf{x} + \mathbf{x}^T \mathbf{M}_2 \mathbf{x} = \mathbf{x}^T (\mathbf{M}_1 + \mathbf{M}_2) \mathbf{x} = \mathbf{x}^T \mathbf{M} \mathbf{x}.$$

This proves that $\mathbf{M}$ is half positive semi-definite.  □

**Theorem 2 ($2 \times 2$ half positive semi-definiteness)** *A sufficient and necessary condition for half positive semi-definiteness of a $2 \times 2$ matrix is that it is a sum of a positive semi-definite and a non-negative matrix.*

**Proof:**  It is easy to prove that for any quadratic form, there exists an equivalent symmetric form, so let $\mathbf{M}$ be a symmetric $2 \times 2$ matrix of elements $a, b, c$

$$\begin{bmatrix} a & c \\ c & b \end{bmatrix}.$$

If $\mathbf{M}$ is half positive semi-definite, we have for any non-negative vector $[x \ y]$

$$[x \ y] \begin{bmatrix} a & c \\ c & b \end{bmatrix} \begin{bmatrix} x \\ y \end{bmatrix} \geq 0.$$

This is equivalent to

$$ax^2 + by^2 + 2cxy \geq 0. \quad (1)$$

It is easy to prove that both $a$ and $b$ have to be non-negative (consider vectors $[x \ 0]$ and $[0 \ y]$ respectively). Now, we distinguish two cases: **(1)** if $c \geq 0$, then $\mathbf{M}$ is non-negative; **(2)** if $c < 0$, then either $a > 0$ or $b > 0$ (note that vector $\mathbf{x} = [1 \ 1]$ yields $a + b + 2c \geq 0$, which given $c < 0$ implies $a + b > 0$). If $a > 0$, then we apply vector $\mathbf{x} = [b \ -c]$ and if $b > 0$, then we apply vector $\mathbf{x} = [-c \ a]$. In both cases, we obtain $ab - c^2 \geq 0$, which is satisfied if and only if $\mathbf{M}$ is positive semi-definite.

We have shown that if a $2 \times 2$ matrix is half positive semi-definite, then it is either non-negative (case 1) or it is positive semi-definite (case 2). This condition is actually stronger than the matrix being a sum of a non-negative and a positive semi-definite matrices. Such strong form of the condition does not hold for $3 \times 3$ matrices.  □

**Theorem 3 (intercausal reasoning)** *Let $a$, $b$, and $x$ be direct predecessors of $c$ such that $a$ and $b$ are conditionally independent* (see Figure 3). *A sufficient and necessary condition for $S^-(a,b)$ on observation of $c_0$ is negative product synergy, $X^-(\{a,b\}, c_0)$.*

**Proof:**  By the definition of qualitative influence we have

$$S^-(ab) \Leftrightarrow \forall_i \forall_{b_1 > b_2} \\ Pr(a > a_i | b_1 c_0) \leq Pr(a > a_i | b_2 c_0). \quad (2)$$

This is equivalent to

$$\forall_i \forall_{b_1 > b_2} \sum_{j=0}^{i-1} [Pr(a_j | b_1 c_0) - Pr(a_j | b_2 c_0)] \leq 0.$$

Expansion of both components by Bayes theorem

$$Pr(a_j | b.c_0) = \frac{Pr(c_0 | a_j b.) Pr(a_j)}{\sum_{k=0}^{n_a} Pr(c_0 | a_k b.) Pr(a_k)},$$

and subsequent simplification yields

$\forall_i \forall_{b_1 > b_2}$

$(\sum_{j=0}^{i-1} \sum_{k=0}^{n_a} Pr(a_j) Pr(a_k) ( Pr(c_0|a_j b_1) Pr(c_0|a_k b_2)$
$- Pr(c_0|a_k b_1) Pr(c_0|a_j b_2) ) ) /$
$\left( \sum_{k=0}^{n_a} Pr(c_0|a_k b_1) Pr(a_k) \sum_{k=0}^{n_a} Pr(c_0|a_k b_2) Pr(a_k) \right) \leq 0.$

We multiply both sides by the denominator and, for the sake of brevity, introduce term $A$ defined as follows

$A_{mn} = Pr(c_0|a_m b_1) \; Pr(c_0|a_n b_2)$
$\qquad\qquad - Pr(c_0|a_n b_1) Pr(c_0|a_m b_2).$

It is straightforward to verify that $\forall_m [A_{mm} = 0]$ and $\forall_{m \neq n} [A_{mn} = -A_{nm}]$. Taking this into consideration, we refine the summation indices, obtaining

$$\forall_i \forall_{b_1 > b_2} \; \sum_{j=0}^{i-1} \sum_{k=i}^{n_a} Pr(a_j) Pr(a_k) A_{jk} \leq 0. \qquad (3)$$

The sufficient and necessary condition for the above to hold for any distribution of $a$ is

$$\forall_{j < k} \; A_{jk} \leq 0.$$

Note here that $j < k$ and we can rewrite this inequality as

$$\forall_{a_1 > a_2} \; A_{12} \leq 0.$$

As $\forall_i [Pr(a_i) \geq 0]$, sufficiency follows directly from (3). We prove the necessity by contradiction. Suppose that for some $b_1 > b_2$ there exist such $a_1 > a_2$ that $A_{12} > 0$. Consider a distribution of $a$ in which $Pr(a_1) > 0$, $Pr(a_2) > 0$, and $Pr(a_1) + Pr(a_2) = 1$. By axioms of probability theory $\forall_{m \neq 1, m \neq 2} [Pr(a_m) = 0]$, which reduces (3) to

$$Pr(a_1) Pr(a_2) A_{12} \leq 0.$$

This implies that $A_{12}$ is not positive, which contradicts the assumption.

We have proven that the sufficient and necessary condition for (2) is

$\forall_{a_1 > a_2} \forall_{b_1 > b_2}$
$\quad Pr(c_0|a_1 b_1) Pr(c_0|a_2 b_2)$
$\quad - Pr(c_0|a_2 b_1) Pr(c_0|a_1 b_2) \leq 0. \qquad (4)$

Note that this condition is equivalent to product synergy I if $c$ has no other predecessors than $a$ and $b$. In order to express this result in terms of the conditional distribution of $c$ given all its immediate predecessors, we introduce $x$ into (4).

$\forall_{a_1 > a_2} \forall_{b_1 > b_2}$

$\quad \sum_{m=0}^{n_x} Pr(c_0|a_1 b_1 x_m) Pr(x_m)$

$\quad \sum_{n=0}^{n_x} Pr(c_0|a_2 b_2 x_n) Pr(x_n)$

$\quad - \sum_{p=0}^{n_x} Pr(c_0|a_2 b_1 x_p) Pr(x_p)$

$\quad \sum_{q=0}^{n_x} Pr(c_0|a_1 b_2 x_q) Pr(x_q) \leq 0.$

After rearranging the summation operators, we get

$\forall_{a_1 > a_2} \forall_{b_1 > b_2} \; \sum_{m=0}^{n_x} \sum_{n=0}^{n_x} Pr(x_m) Pr(x_n)$
$\quad ( Pr(c_0|a_1 b_1 x_m) Pr(c_0|a_2 b_2 x_n)$
$\quad - Pr(c_0|a_2 b_1 x_m) Pr(c_0|a_1 b_2 x_n) ) \leq 0,$

which is equal to

$$\forall_{a_1 > a_2} \forall_{b_1 > b_2} \; \sum_{m=0}^{n_x} \sum_{n=0}^{n_x} Pr(x_m) Pr(x_n) D_{mn} \leq 0. \qquad (5)$$

This can be written in matrix notation as

$$\forall_{a_1 > a_2} \forall_{b_1 > b_2} \; \mathbf{p}^T \mathbf{D} \mathbf{p} \leq 0. \qquad (6)$$

where $\mathbf{p}$ is a vector of probabilities of various outcomes of $x$ ($p_i = Pr(x_i)$), and $\mathbf{D}$ is a square matrix with elements $D_{mn}$. Inequality (6) will hold for any vector of probabilities $\mathbf{p}$ if and only if $\mathbf{D}$ is half negative semi-definite, which is exactly the condition for the negative product synergy II.   □

**Theorem 4 (intercausal reasoning)** *Let $a$, $b$, and $x$ be direct predecessors of $c$, and $c$ be a direct predecessor of $d$ in a network. Let $c$ be binary. Let there be no direct links from $a$ or $b$ to $d$ (see Figure 4). Let $X^{\delta_1}(\{a, b\}, C)$, $X^{\delta_2}(\{a, b\}, \overline{C})$, $Y^{\delta_3}(\{a, b\}, c)$, and $S^{\delta_4}(c, d)$.*
*If $\delta_4 = +$ and $\delta_1 = \delta_3$, then $S^{\delta_1}(a, b)$ holds in the network with $D$ observed. If $\delta_4 = -$ and $\delta_2 \neq \delta_3$, then $S^{\delta_2}(a, b)$ holds in the network with $D$ observed.*

**Proof:** Let $n_a$, $n_c$, and $n_x$ denote the number of possible values of $a$, $c$, and $x$ respectively. By the definition of qualitative influence

$$S^-(ab) \Leftrightarrow \forall_i \; Pr(a > a_i|b_1 d_0) \leq Pr(a > a_i|b_2 d_0).$$

This is equivalent to

$$\forall_i \; \sum_{j=0}^{i-1} [Pr(a_j|b_1 d_0) - Pr(a_j|b_2 d_0)] \leq 0.$$

Expansion of both components by Bayes theorem

$Pr(a_j|b.d_0) = \frac{Pr(d_0|a_j b.) Pr(a_j b.)}{Pr(d_0|b.) Pr(b.)}$
$\qquad\qquad = \frac{Pr(d_0|a_j b.) Pr(a_j)}{Pr(d_0|b.)}$

and simplification yields

$\forall_i \; \sum_{j=0}^{i-1} Pr(a_j)$
$\quad \frac{Pr(d_0|a_j b_1) Pr(d_0|b_2) - Pr(d_0|a_j b_2) Pr(d_0|b_1)}{Pr(d_0|b_1) Pr(d_0|b_2)} \leq 0.$

Multiplying both sides of the inequality by the denominator, which does not depend on the summation index and is positive, yields

$$\forall_i \sum_{j=0}^{i-1} Pr(a_j)( Pr(d_0|a_jb_1)Pr(d_0|b_2) - Pr(d_0|a_jb_2)Pr(d_0|b_1) ) \leq 0.$$

We expand the formulas for $Pr(d_0)$ using

$$Pr(d_0|a_jb.) = \sum_{k=0}^{n_c} Pr(d_0|c_k)Pr(c_k|a_jb.)$$

and

$$Pr(d_0|b.) = \sum_{m=0}^{n_c} Pr(d_0|c_m) \sum_{n=0}^{n_a} Pr(c_m|a_nb.)Pr(a_n),$$

which, after rearranging the summation terms, yields

$$\forall_i \sum_{j=0}^{i-1}\sum_{n=0}^{n_a} Pr(a_j)Pr(a_n) \sum_{k=0}^{n_c}\sum_{m=0}^{n_c} Pr(d_0|c_k)Pr(d_0|c_m)$$
$$( Pr(c_k|a_jb_1)Pr(c_m|a_nb_2) - Pr(c_k|a_jb_2)Pr(c_m|a_nb_1) ) \leq 0. \quad (7)$$

For a binary $c$ (i.e., $n_c = 2$, $c_0 = C$, $c_1 = \overline{C}$), (7) takes the following form

$$\forall_i \quad \sum_{j=0}^{i-1}\sum_{k=0}^{n_a} Pr(a_j)Pr(a_k)$$
$$Pr(d_0|C)Pr(d_0|C)( Pr(C|a_jb_1)Pr(C|a_kb_2) - Pr(C|a_jb_2)Pr(C|a_kb_1) )$$
$$+ Pr(d_0|C)Pr(d_0|\overline{C})( Pr(C|a_jb_1)Pr(\overline{C}|a_kb_2) - Pr(C|a_jb_2)Pr(\overline{C}|a_kb_1) )$$
$$+ Pr(d_0|\overline{C})Pr(d_0|C)( Pr(\overline{C}|a_jb_1)Pr(C|a_kb_2) - Pr(\overline{C}|a_jb_2)Pr(C|a_kb_1) )$$
$$+ Pr(d_0|\overline{C})Pr(d_0|\overline{C})( Pr(\overline{C}|a_jb_1)Pr(\overline{C}|a_kb_2) - Pr(\overline{C}|a_jb_2)Pr(\overline{C}|a_kb_1) ) \leq 0.$$

We divide both sides twice by $Pr(d|\overline{C})$ and substitute $\lambda$ for the likelihood ratio $Pr(d|C)/Pr(d|\overline{C})$. Rearrangement and simplification yields

$$\forall_i \sum_{j=0}^{i-1}\sum_{k=0}^{n_a} Pr(a_j)Pr(a_k)(\lambda - 1)$$
$$(( \lambda( Pr(C|a_jb_1)Pr(C|a_kb_2) - Pr(C|a_jb_2)Pr(C|a_kb_1) )$$
$$- ( Pr(\overline{C}|a_jb_1)Pr(\overline{C}|a_kb_2) - Pr(\overline{C}|a_jb_2)Pr(\overline{C}|a_kb_1) ) ) \leq 0.$$

For the sake of brevity we introduce terms $C_{mn}$ and $\overline{C}_{mn}$ defined as follows

$$C_{mn} = Pr(C|a_mb_1)Pr(C|a_nb_2) - Pr(C|a_mb_2)Pr(C|a_nb_1)$$
$$\overline{C}_{mn} = Pr(\overline{C}|a_mb_1)Pr(\overline{C}|a_nb_2) - Pr(\overline{C}|a_mb_2)Pr(\overline{C}|a_nb_1).$$

It is straightforward to verify that $\forall_m [C_{mm} = 0]$ and $\forall_{m \neq n} [C_{mn} = -C_{nm}]$. Analogous conditions are valid for $\overline{C}_{mn}$. Taking this into consideration, we refine the summation indices, obtaining

$$\forall_i \sum_{j=0}^{i-1}\sum_{k=i}^{n_a} Pr(a_j)Pr(a_k)(\lambda - 1)(\lambda C_{jk} - \overline{C}_{jk}) \leq 0. \quad (8)$$

Note here that $j < k$. The sufficient and necessary condition for the above to hold for any distribution of $a$ is

$$\forall_{j<k} \quad (\lambda - 1)(\lambda C_{jk} - \overline{C}_{jk}) \leq 0. \quad (9)$$

As $\forall_i [Pr(a_i) \geq 0]$, sufficiency follows directly from (8). We prove the necessity by contradiction. Suppose there exist $j$ and $k$ such that $(\lambda - 1)(\lambda C_{jk} - \overline{C}_{jk}) > 0$. Consider a distribution of $a$ in which $Pr(a_j) > 0$, $Pr(a_k) > 0$, and $Pr(a_j) + Pr(a_k) = 1$. By axioms of probability theory $\forall_{m \neq j, m \neq k} [Pr(a_m) = 0]$, which reduces (8) to

$$Pr(a_j)Pr(a_k)(\lambda - 1)(\lambda C_{jk} - \overline{C}_{jk}) \leq 0.$$

This implies that $(\lambda - 1)(\lambda C_{jk} - \overline{C}_{jk})$ is negative, which contradicts the assumption.

We have proven that the sufficient and necessary condition for (8) is

$$\forall_{j<k}$$
$$(\lambda - 1)(( \lambda( Pr(C|a_jb_1)Pr(C|a_kb_2) - Pr(C|a_jb_2)Pr(C|a_kb_1) ) \quad (10)$$
$$- ( Pr(\overline{C}|a_jb_1)Pr(\overline{C}|a_kb_2) - Pr(\overline{C}|a_jb_2)Pr(\overline{C}|a_kb_1) ) ) \leq 0.$$

Substituting $Pr(\overline{C}) = 1 - Pr(C)$ in (11), and simplifying yields an equivalent formula

$$\forall_{j<k} \quad (\lambda - 1)$$
$$(( \lambda - 1)( Pr(C|a_jb_1)Pr(C|a_kb_2) - Pr(C|a_jb_2)Pr(C|a_kb_1) ) \quad (11)$$
$$+ Pr(C|a_jb_1) - Pr(C|a_kb_1)$$
$$+ Pr(C|a_jb_2) - Pr(C|a_kb_2) ) \leq 0.$$

In order to express both results in terms of the conditional distribution of $c$ given all its immediate predecessors, we introduce $x$ into (11)

$$\forall_{j<k} \quad (\lambda - 1)$$
$$( \lambda \sum_{m=0}^{n_x}\sum_{n=0}^{n_x} Pr(x_m)Pr(x_n)$$
$$( Pr(C|a_jb_1x_m)Pr(C|a_kb_2x_n) - Pr(C|a_jb_2x_m)Pr(C|a_kb_1x_n) )$$
$$- \sum_{m=0}^{n_x}\sum_{n=0}^{n_x} Pr(x_m)Pr(x_n)$$
$$( Pr(\overline{C}|a_jb_1x_m)Pr(\overline{C}|a_kb_2x_n) - Pr(\overline{C}|a_jb_2x_m)Pr(\overline{C}|a_kb_1x_n) ) ) \leq 0.$$

The above can be written using matrix notation as

$$\forall_{j<k} \quad (\lambda - 1)\left(\lambda \mathbf{p}^T \mathbf{D} \mathbf{p} - \mathbf{p}^T \overline{\mathbf{D}} \mathbf{p}\right) \leq 0 \quad (12)$$

where $\mathbf{p}$ is a vector of probabilities of various outcomes of $x$ ($p_i = Pr(x_i)$), $\mathbf{D}$ and $\overline{\mathbf{D}}$ are square matrices with elements $D_{mn}$ for $c = C$ and $c = \overline{C}$ respectively.

Replacement of the matrix expressions by the formulas used for computing the value of product synergies from the numerical distribution (we will denote the fact that they are formulas and not the synergies by enclosing them in straight brackets, e.g., $|X^{\delta_2}(\{a,b\},C)|$) yields

$$\forall_{j<k} \quad (\lambda - 1)$$
$$\left(\lambda \left|X^{\delta_1}(\{a,b\},C)\right| - \left|X^{\delta_2}(\{a,b\},\overline{C})\right|\right) \leq 0. \quad (13)$$

A similar procedure with respect to (12) yields

$$\forall_{j<k} \quad (\lambda - 1)$$
$$( \lambda \sum_{m=0}^{n_x} \sum_{n=0}^{n_x} Pr(x_m) Pr(x_n)$$
$$( Pr(C|a_j b_1 x_m) Pr(C|a_k b_2 x_n)$$
$$- Pr(C|a_j b_2 x_m) Pr(C|a_k b_1 x_n) )$$
$$- \sum_{n=0}^{n_x} Pr(x_n)$$
$$( Pr(C|a_j b_1 x_n) - Pr(C|a_k b_1 x_n)$$
$$+ Pr(C|a_j b_2 x_n) - Pr(C|a_k b_2 x_n) ) ) \leq 0.$$

and with the expressions for product and additive synergy

$$\forall_{j<k} \quad (\lambda - 1) \quad (14)$$
$$\left((\lambda - 1)\left|X^{\delta_1}(\{a,b\},C)\right| + \left|Y^{\delta_3}(\{a,b\},c)\right|\right) \leq 0.$$

It is clear that the formulas (13) and (15) have at most two zero points for different values of evidential support $\lambda$. One of this points is $\lambda = 1$ and the other can be theoretically anywhere (including $\lambda < 0$, which as $0 \leq \lambda < \infty$ means that there is only one zero point for the possible values of $\lambda$).

For $\lambda = 0$, the condition for intercausal reasoning with indirect support transforms into $X^{\delta_2}(\{a,b\},\overline{C})$. We verified also that the complete formula (before reducing the denominator) reduces to $X^{\delta_1}(\{a,b\},C)$ as $\lambda \to \infty$. It is easy to verify that that if $\delta_1 \neq \delta_2$, then the second zero point cannot be in the interval $0 \leq \lambda < \infty$. In such case, the sign of intercausal inference is unambiguous, and equal for $\delta_2$ if $0 \leq \lambda < 1$ and equal for $\delta_1$ if $1 < \lambda < \infty$. In case the signs of the two product synergies are equal, the additive synergy determines the interval in which the second zero point falls and determines the sign of the remaining interval unambiguously. □

## Acknowledgments

Funding for this research was provided by the Rockwell International Science Center. We thank Professor Richard Duffin for suggesting the term *half positive semi-definiteness*. Professors Victor Mizel and Juan Schaffer were the first to note the conditions for half positive semi-definiteness. Professor Schaffer's suggestion improved our proof of Theorem 2. Anonymous reviewers provided useful remarks.